\documentclass[runningheads,a4paper]{llncs}
\usepackage{graphicx} % Required for inserting images
\usepackage{amsmath}
\usepackage{amsfonts}
\usepackage{bm}
\usepackage{booktabs}
\usepackage{adjustbox}
\usepackage{todonotes}
\usepackage{hyperref}
\usepackage{cite}
\usepackage{multirow}
\usepackage{appendix}
\usepackage{xr-hyper}
\newcommand*\dif{\mathop{}\!\mathrm{d}}

\title{Expert-Guided POMDP Learning for
Data-Efficient Modeling in Healthcare}

\titlerunning{Expert-Guided POMDP Learning}
\author{Marco Locatelli\inst{1}\orcidID{0009-0000-2220-2023} \and
Arjen Hommersom\inst{2}\orcidID{0000-0003-0125-1680}
\and
Roberto Clemens Cerioli\inst{1}\orcidID{0009-0003-8063-1311} 
\and
Daniela Besozzi\inst{1,3,4}\orcidID{0000-0001-5532-3059}
\and
Fabio Stella\inst{1,3,4}\orcidID{0000-0002-1394-0507}}
\authorrunning{M. Locatelli et al.}

\institute{Department of Informatics, Systems and Communication, University of Milano-Bicocca, Milan, Italy
\and
Department of Computer Science, Open University, P.O. Box 2960, 6401DL Heerlen, The Netherlands
\and 
Bicocca Bioinformatics Biostatistics and Bioimaging Centre - B4, University of Milano-Bicocca, Vedano al Lambro (MB), Italy 
\and
BReCHS – Bicocca Research Centre in Health Services, University of Milano-Bicocca, Milan, Italy}

\begin{document}

\maketitle

\begin{abstract} Learning the parameters of Partially Observable Markov Decision Processes (POMDPs) from  limited data is a significant challenge. We introduce the Fuzzy-MAP EM algorithm, a novel approach that incorporates expert knowledge into the parameter estimation process by enriching the Expectation–Maximization (EM) framework with fuzzy pseudo-counts derived from an expert-defined fuzzy model. This integration naturally reformulates the problem as a Maximum A Posteriori (MAP) estimation, effectively guiding learning in environments with limited data. In synthetic medical simulations, our method consistently outperforms the standard EM algorithm under both low-data and high-noise conditions. Furthermore, a case study on Myasthenia Gravis illustrates the ability of the Fuzzy-MAP EM algorithm to recover a clinically coherent POMDP, demonstrating its potential as a practical tool for data-efficient modeling in healthcare.
\end{abstract}

\section{Introduction} Sequential decision-making poses a critical challenge across numerous real-world domains, particularly in healthcare, where diagnostic and treatment decisions must be made over time in response to patients' dynamic conditions. This complexity is further amplified in the context of rare diseases, which are marked by small, heterogeneous patients populations, limited clinical data, and an incomplete understanding of underlying disease mechanisms.

%Dynamic Treatment Regimes (DTRs) have become a widely utilized framework for managing chronic diseases, offering data-driven decision rules that guide treatment choices over time based on a patient’s evolving clinical profile. DTRs are particularly effective when the decision process can be reliably supported by a set of observable markers that accurately reflect the patient's underlying condition. However, DTRs, face limitations in more complex settings, such as rare diseases, where critical aspects of disease progression may not be fully captured by observable features. A further challenge lies in the integration of prior knowledge, such as expert insights and known biological mechanisms—into the formulation of DTRs, which are typically data-driven and may not readily accommodate such information.

Partially Observable Markov Decision Processes (POMDPs) provide a general framework for making sequential decisions in environments where the true state of the system is uncertain or hidden \cite{cassandra1994acting}. POMDPs maintain a probabilistic belief over latent states and use this information to inform decision-making. This enables policies to evolve over time by combining current observations with probabilistic estimates of disease progression, making POMDPs particularly well-suited for rare diseases, which are characterized by limited data availability and substantial uncertainty \cite{hauskrecht2000planning}.
Despite their theoretical advantages, POMDPs face practical challenges: {\em i}) exact solution methods scale poorly with state and observation complexity, and {\em ii}) learning parameters under data paucity is difficult. 

In this work a novel Fuzzy-MAP EM algorithm is designed to tackle the second challenge by integrating medical knowledge into a POMDP model. Our approach leverages expert knowledge of disease dynamics, encoded within a fuzzy modeling framework, to inform a Maximum A Posteriori (MAP) estimation algorithm. 
The fuzzy model generates \textit{fuzzy pseudo-counts} that serve as informative priors, thereby regularizing parameter updates and enhancing the robustness of the estimation process.
To the best of our knowledge, constructing a prior using fuzzy rules for the M-step of the EM algorithm represents a novel methodological contribution.
Specifically, the central contributions of this study are:
\begin{itemize}
    \item The introduction of the Fuzzy-MAP EM algorithm, a novel method for integrating expert knowledge into POMDP parameter estimation by augmenting the M-step of the EM algorithm with \textit{fuzzy pseudo-counts}.
    \item A demonstration, via synthetic medical simulations, that the proposed approach consistently outperforms the standard EM algorithm, particularly in challenging low-data and high-noise environments.
    \item  A validation of the algorithm's practical utility through a real-world case study on Myasthenia Gravis, a rare autoimmune disorder that impairs communication between nerves and muscles. This study demonstrates the possibility to learn a clinically coherent POMDP from expert knowledge alone.
\end{itemize}

%The effectiveness of the proposed algorithm is first demonstrated in a synthetic medical environment, where it yields substantial improvements in the precision of the learned transition model under conditions of limited data and high noise. Subsequently, a practical case study is presented on the management of Myasthenia Gravis (MG), a rare autoimmune disorder that impairs communication between nerves and muscles. In this study, the underlying fuzzy model of disease progression was constructed entirely by combining clinical literature and expert knowledge. This illustrates the capability of our method to establish a robust decision-making framework even in scenarios where patient data is initially unavailable.

This paper begins by reviewing the foundational concepts of POMDPs, the EM algorithm, and Takagi-Sugeno fuzzy models in Section~\ref{sec:preliminaries}. Section~\ref{sec:Fuzzy-MAP EM} introduces our novel Fuzzy-MAP EM algorithm, and describes how expert knowledge is integrated through {\em fuzzy pseudo-counts}. In Section~\ref{sec:experiments}, we evaluate the algorithm’s performance in a simulated medical environment characterized by limited and noisy data. Section~\ref{sec:miastehenia} presents a real-world case study focused on the management of Myasthenia Gravis, a rare autoimmune disease. Finally, Section~\ref{sec:conclusion}
concludes the paper and outlines directions for future research.

\section{Preliminaries}
\label{sec:preliminaries}
%\todo[color=pink, textcolor=black]{What about naming this section as Methods? (DB)}
%\todo[color=white, textcolor=red]{These are methods propedeutic to our contributions, does it fit the same to call them methods? (FS)}

%In this section, we introduce the fundamental definitions and notations related to Partially Observable Markov Decision Processes (POMDPs) and Takagi-Sugeno fuzzy models.

\subsection{Partially Observable Markov Decision Processes}
A Partially Observable Markov Decision Processes (POMDP) \cite{cassandra1994acting} is a mathematical framework for modeling an agent acting in a stochastic, partially observable environment.

A POMDP is formally defined as a tuple $\langle S, A, \Omega, T, O, R \rangle$, where:
\begin{itemize}
    \item $S$ is a finite set of hidden states.
    \item $A$ is a finite set of actions.
    \item $\Omega$ is a continuous observation space, $\Omega \subseteq \bbbr^d$. The vector of observations at time $t$ is represented by $o_t \in \Omega$,  where $o_{t,j}$ denotes the $j$-th component.
    %with $o_t \in \Omega \subseteq \mathbb{R}^d$ being an observation vector at time $t$.
    \item  $T(s, a, s') = P(s_{t+1} = s' \mid s_t = s, a_t = a)$ is the state transition probability function, which defines the probability of transitioning from state $s$ to state $s'$ after taking action $a$.
    \item $O(o \mid s) = p(o_t = o \mid s_t = s)$ is the observation probability density function, which defines the probability of observing $o$ when the system is in state $s$.
    \item $R(s, a, s')$ is the reward function, which specifies the immediate reward for transitioning from $s$ to $s'$ via action $a$. While crucial for control, the reward function is not involved in the parameter learning problem addressed here.
\end{itemize}

\subsection{Expectation-Maximization for POMDPs}
The Expectation-Maximization (EM) algorithm \cite{dempster1977maximum} is an iterative algorithm for finding the maximum likelihood estimates of parameters in probabilistic models with latent variables. For POMPDs the latent variables are the unobserved state trajectories $s_{1:T_i}^{(i)}$, where $(i)$ indexes a specific patient trajectory in the dataset and $1:T_i$ denote the sequence of time steps for that trajectory. EM treats each full state sequence as missing data and marginalizes over it. 

Given a dataset $\mathcal{D} = \{ a_{1:T_i}^{(i)}, o_{1:T_i}^{(i)} \}_{i
=1}^N$ consisting of $N$ action-observation trajectories, the goal is to find the parameters $\theta$ that maximize the log-likelihood of the data:
\begin{equation}
    \mathcal{L}(\theta) = \sum_{i=1}^N \log P(o_{i:T_i}^{(i)} \mid a_{i:T_i-1}^{(i)} ; \theta).
\end{equation}

The EM algorithm solves this problem 
%in an iterative fashion 
by alternating between two steps:
%\vspace{-0.5cm}
\paragraph{E-Step (Expectation).} Given the parameters estimate $\theta^k$ at the $k\text{-}th$ EM iteration, this step computes the posterior probability of the latent sequences using the forward-backward algorithm \cite{baum1970maximization}. In  detail, the computed quantities are:
%the following two:
\begin{enumerate}
    \item \textbf{State occupancy probability:} the probability of being in state $s$ at time $t$: $ \gamma_t(s) = P(s_t = s \mid o_{i:T_i}, a_{i:T_i-1}; \theta^k)$.
    \item \textbf{State transition probability:} the joint probability of transitioning from state $s$ to state $s'$ at time $t$: $\xi_t(s, s') = P(s_t = s, s_{t+1} = s' \mid o_{i:T_i}, a_{i:T_i-1}; \theta^k)$.
\end{enumerate}
%\vspace{-0.5cm}
\paragraph{M-Step (Maximization).} This step updates the model parameters from $\theta^{k}$ to $\theta^{k+1}$ by maximizing the expected complete-data log-likelihood. This is obtained by using the probabilities values computed by the {\em E-step} as expected counts.
\begin{itemize}
    \item \textbf{Transition model update}: $T^{k+1}(s, a, s') = \frac{\sum_{i=1}^N \sum_{t=1}^{T_i-1} \mathbb{I}(a_t^{(i)} = a) \xi_t(s, s') }{\sum_{i=1}^N \sum_{t=1}^{T_i-1} \mathbb{I}(a_t^{(i)} = a) \gamma_t(s)} = \frac{\hat{N}_T(s, a, s')}{\sum_{s'' \in S} \hat{N}_T(s, a, s'') }$,
    where $\mathbb{I}(\cdot)$ is the indicator function.

    \item \textbf{Observation model update} (for a multivariate normal distribution): 
    \begin{align*}
        \mu_s^{(k+1)} &= \frac{\sum_{i=1}^N \sum_{t=1}^{T_i} \gamma_t^{(i)}(s) o_t^{(i)}}{\sum_{i=1}^N \sum_{t=1}^{T_i} \gamma_t^{(i)}(s)} = \frac{\hat{S}_O(s)}{\hat{N}_O(s)}\\
        \\
        \Sigma_s^{(k+1)} &= \frac{\sum_{i=1}^N \sum_{t=1}^{T_i} \gamma_t^{(i)}(s) (o_t^{(i)} - \mu_s^{(k+1)})(o_t^{(i)} - \mu_s^{(k+1)})^{\mathrm T}}{\sum_{i=1}^N \sum_{t=1}^{T_i} \gamma_t^{(i)}(s)}\\
        &= \frac{\hat{S}_{O,2}(s)}{\hat{N}_O(s)} -\mu_{s'}^{(k+1)} \mu_{s'}^{(k+1)\,\mathrm T}
    \end{align*}
\end{itemize}

\subsection{Takagi-Sugeno Fuzzy Models}
To integrate expert knowledge into the learning process, we employ a Type-1 Takagi-Sugeno fuzzy model \cite{mehran2008takagi}. This model predicts the evolution of the system's observations.
A Type-1 Takagi-Sugeno fuzzy model consists of:
\begin{itemize}
    \item \textbf{Membership Functions}: Each input variable (e.g., an observation component) is associated with a set of linguistic terms (LTs) such as ``low", ``medium" and ``high". A membership function $\mu_{LT}(x) \in [0, 1]$ quantifies the degree to which a crisp input value belongs to a linguistic concept.
%\todo[color=pink, textcolor=black]{Formally, low/high are generally referred to as linguistic terms, whereas the linguistic variable is the name of the input variable itself. (DB)}
    \item \textbf{Fuzzy Rules}: The system's behavior is described by a set $R$ of IF-THEN rules. A typical fuzzy rule $r \in R$ has the form:
    \begin{center}
        IF $o_{t,1}$ is $LT_1$ AND ... AND $a_t$ is $LT_a$ THEN $o^*_{t+1,j} = f_{r,j}(o_t, a_t)$,
    \end{center}
    where $f_{r,j}$ is a function (typically linear) of the inputs for rule $r$.

    \item \textbf{Firing Strength}: The firing strength of a rule's antecedent, $\mu_{r,ant}(o_t, a_t)$, is calculated using a t-norm operator ($\top$) on the membership values of the inputs:
    \begin{equation}
        \mu_{r,ant}(o_t, a_t) = \top(\mu_{LT_{r,1}}(o_{t,1}), ..., \mu_{LT_{r,a}}(a_t)).
    \end{equation}

    \item \textbf{Inference}: The model's final output, $\hat{o}_{t+1}$, is a weighted average of the consequents of all rules, where the weights are the firing strengths:
    \begin{equation}
        \hat{o}_{t+1} = \frac{\sum_{r} \mu_{r,ant}(o_t, a_t) \cdot f_r(o_t, a_t)}{\sum_{r} \mu_{r,ant}(o_t, a_t)}.
    \end{equation}
    This provides a prediction for the next observation, denoted as $\hat{o}_{t+1} = f_{\text{fuzzy}}(o_t, a_t)$.
\end{itemize}

\section{A Fuzzy-MAP Expectation-Maximization Algorithm}
\label{sec:Fuzzy-MAP EM}
The core of our method involves augmenting the M-step of the Expectation-Maximization (EM) algorithm with priors derived from a fuzzy model defined by domain experts. This approach transforms the standard Maximum Likelihood Estimation into a Maximum A Posteriori (MAP) estimation \cite{gauvain1994maximum}, where the fuzzy model serves as a source of prior knowledge. Integration is achieved by generating fuzzy pseudo-counts, which are incorporated into the empirical counts obtained during the E-step, thereby guiding the parameter updates.
 %This prior acts as a regularizer, guiding the solution towards expert heuristics, a crucial function when data is insufficient.

\subsection{Fuzzy Pseudo-Counts}
We generate separate pseudo-counts for both the transition model and the observation model. The process involves evaluating each fuzzy rule's relevance and prediction in the context of the current POMDP parameters estimate $\theta^k$.

The formulation of  fuzzy transition pseudo-counts has two key components:
\begin{itemize}
    %\item  The \textbf{Matching Antecedent Score}, $MatchAnt(s, a, r)$, quantifies how strongly the antecedent of a fuzzy rule $r$ is activated when the system is in a latent state $s$ and action $a$ is taken. 
    \item \textbf{Matching Antecedent Score}, denoted by $MatchAnt(s, a, r)$, represents the expected firing strength with respect to the observation distribution associated with state $s$. Conceptually, it quantifies how strongly the antecedent of a fuzzy rule $r$ is activated when the system is in a latent state $s$ and action $a$ is taken:
    \begin{equation}
        MatchAnt(s, a, r)= \int_{o \in \Omega} \mu_{r,ant}(o, a) \: O(o \mid s; \theta^k) \dif o.
        \label{eq:matchant}
    \end{equation}
    %\item  The \textbf{Consequent Likelihood} quantifies the compatibility of a fuzzy rule's consequent with the observation model of a potential future state $s'$.
    \item \textbf{Consequent Likelihood} evaluates how likely the consequent of a fuzzy rule is to have been generated by the subsequent state $s'$. It assesses the plausibility of a fuzzy rule's prediction within the context of the POMDP's state dynamics:
    \begin{equation}
                 L(Y_r^*(s,a) \mid s'; \theta^{(k)}) = O(Y_r^*(s,a) \mid s'; \theta^{(k)}),
                 \label{eq:cons_like}
    \end{equation}
    where
    \begin{equation}
        Y_r^*(s,a) = \mathbb{E}_{o \sim O(\cdot \mid s; \theta^{(k)})}\big[f_r(o,a)\big] = \int f_r(o,a) \cdot O(o\mid s; \theta^{(k)}) \dif o.\\
    \end{equation}
\end{itemize}
   
\begin{definition}
    The \textbf{fuzzy transition pseudo-counts} for a transition ($s, a, s'$) is the sum over all fuzzy rules of the product of the antecedent match (Eq. \ref{eq:matchant}) and the consequent likelihood (Eq. \ref{eq:cons_like}):
\begin{equation}
        N_T^{\text{fuzzy}}(s, a, s') = \sum_{r=1}^R \text{MatchAnt}(s, a, r) \times L(Y_r^*(s, a) \mid s'; \theta^{(k)}).
    \end{equation}
\end{definition}

In order to compute the fuzzy observation pseudo-counts, it is necessary to introduce the concept of $strength(s, a, s', r)$. The term $strength(s, a, s', r)$ represents the expected rule activation for a specific transition and is defined as:
\begin{equation}
    strength(s, a, s', r) = T^{(k)}(s, a, s') \times MatchAnt(s, a, r).
\end{equation}

\begin{definition}
    The \textbf{fuzzy observation pseudo-counts} are computed for each state $s'$ by marginalizing over all possible antecedent states $s$, actions $a$, and fuzzy rules $r$. The contribution of each rule's consequent, $Y_r^*(s, a)$, is weighted by its corresponding  $\text{strength}(s, a, s', r)$:
    \begin{align}
        N_O^{\text{fuzzy}}(s') &= \sum_{r \in R} \sum_{s \in S} \sum_{a \in A} \text{strength}(s, a, s', r) \\
        S_O^{\text{fuzzy}}(s') &= \sum_{r \in R} \sum_{s \in S} \sum_{a \in A} \text{strength}(s, a, s', r) \times Y_r^*(s, a)\\
        S_{O,2}^{\text{fuzzy}}(s') &= \sum_{r \in R} \sum_{s \in S} \sum_{a \in A} \text{strength}(s, a, s', r) \times (Y_r^*(s, a) \cdot Y_r^*(s, a)^{\mathrm T})
    \end{align}

\end{definition}

\subsection{Fuzzy Pseudo-Counts Augmented M-step}
The M-step is augmented by adding the fuzzy pseudo-counts, weighted by the hyperparameters $\lambda_T$ and $\lambda_O$, to the empirical counts derived from the E-step.
The augmented counts are:
\begin{align}
    \tilde{N}_T(s, a, s') &= \hat{N}_T(s, a, s') + \lambda_T N_T^{\text{fuzzy}}(s, a, s') \\
    \tilde{N}_O(s) &= \hat{N}_O(s) + \lambda_O N_O^{\text{fuzzy}}(s) \\
    \tilde{S}_O(s) &= \hat{S}_O(s) + \lambda_O S_O^{\text{fuzzy}}(s) \\
    \tilde{S}_{O,2}(s) &= \hat{S}_{O,2}(s) + \lambda_O S_{O,2}^{\text{fuzzy}}(s)
\end{align}
The parameter updates then proceed using these augmented counts:
\begin{itemize}
    \item \textbf{Transition Probabilities}:
    \begin{equation}
        T^{(k+1)}(s, a, s') = \frac{\tilde{N}_T(s, a, s')}{\sum_{s'' \in S} \tilde{N}_T(s, a, s'')}
    \end{equation}
    \item \textbf{Observation Parameters} (for each state $s'$):
    \begin{align}
        \mu_{s'}^{(k+1)} &= \frac{\tilde{S}_O(s')}{\tilde{N}_O(s')}, &&&
        \Sigma_{s'}^{(k+1)} &= \frac{\tilde{S}_{O,2}(s')}{\tilde{N}_O(s')} - \mu_{s'}^{(k+1)} \mu_{s'}^{(k+1)\,\mathrm T}
    \end{align}
\end{itemize}

\section{Numerical Experiments on Synthetic Data}
\label{sec:experiments}
The aim of this section is to evaluate the impact of integrating expert knowledge into the parameter estimation process for POMDPs. To achieve this, we adopt a comparative framework, benchmarking the performance of Fuzzy-MAP EM against the standard EM algorithm, which serves as our baseline reference.
%The primary objective of this section is to assess the effectiveness of incorporating expert knowledge into the parameter estimation process for POMDPs. To this end, we adopt a comparative approach, evaluating the performance of the proposed Fuzzy-MAP EM algorithm against the standard EM algorithm, which serves as our baseline.

To assess model performance, we compare the learned parameters against the ground truth of the underlying POMDP. The accuracy of the transition model is quantified using the L1 distance, which measures the total absolute deviation between the estimated and true transition probabilities. For the observation model, we use the Kullback–Leibler (KL) divergence \cite{kullback1951information} to evaluate the difference between the learned and actual observation distributions for each state, capturing how well the model approximates the true generative process.

%Model performance is evaluated by comparing the learned parameters to the ground truth of the underlying POMDP. To quantify the accuracy of the transition model, we use the L1 distance, which captures the absolute error in the estimated transition probabilities. For the observation model, we employ the Kullback–Leibler (KL) divergence to measure the discrepancy between the learned and true observation distributions for each state. 

Since the learning process is unsupervised, the latent states inferred by the model do not possess inherent semantic labels. To assign meaningful labels (e.g., ``Healthy'', ``Sick'', ``Critical''), we manually compared the observed distributions associated with each latent state to the ground-truth distributions, and matched them based on their similarity.

%\subsection{Synthetic Environment}
%\todo[color=pink, textcolor=black]{Sect. 4.1 might be directly merged into Sect. 4, maybe changing the title into 'Numerical Experiments on Synthetic Medical Environment' or 'Fuzzy-MAP EM on Synthetic Medical Environment'. (DB)}

To evaluate our approach, we created a simulated medical environment in which a patient's latent state can be one of the following:  $\{ \text{Healthy}, \text{Sick}, \text{Critical}\}$. 
At each time step $t$, a clinician selects an action from the set $\{ \text{Wait}, \text{Treat}\}$.
Taking the selected action, the clinician observes a two-dimensional vector $o_t = (\text{symptoms}_t, \text{test\_result}_t)$, where each component is a continuous variable taking values within the interval $[0, 1]$.

\begin{figure}[!ht]
    \centering
    \includegraphics[alt={Ground-truth observation distributions}, width=0.95\linewidth]{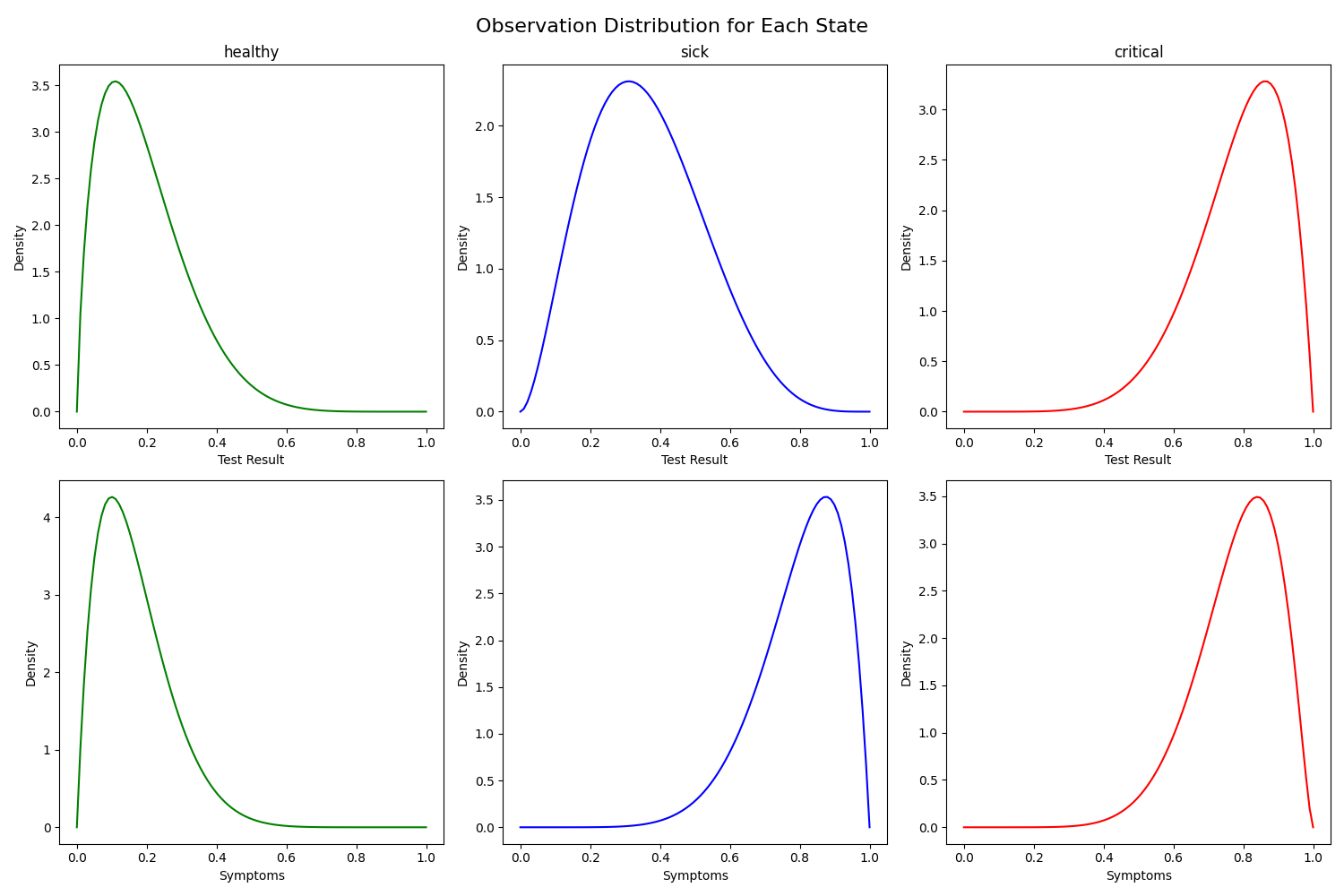}
    \caption{Ground-truth observation distributions for each state.}
    \label{fig:beta_dist}
\end{figure}

The underlying dynamics of this environment are governed by a ground-truth POMDP whose state transition probabilities, $P(s' \mid s,a)$, are given in Table \ref{tab:transitions}. 
For each hidden state, observations are drawn from a corresponding Beta distribution, as depicted in Fig. \ref{fig:beta_dist}.
Throughout the learning phase, the underlying distributions are approximated by multivariate normal distributions.

\begin{table}[!htp]
    \centering
    \caption{Ground-truth transition probabilities $P(s' \mid s,a)$.}
    \label{tab:transitions}
    \begin{tabular}{llccc}
        \toprule
        \textbf{Action} \hspace{0.7cm} & \textbf{Current State} \hspace{0.7cm} & \multicolumn{3}{c}{\textbf{Next State}} \\
        \cmidrule(lr){3-5}
         & & Healthy\hspace{0.4cm} & Sick\hspace{0.4cm} & Critical \\
        \midrule
        \multirow{3}*{Wait} & Healthy & 0.85 & 0.14 & 0.01 \\
             & Sick    & 0.30 & 0.60 & 0.10 \\
             & Critical& 0.05 & 0.01 & 0.94 \\
        \midrule
        \multirow{3}*{Treat} & Healthy & 0.80 & 0.15 & 0.05 \\
              & Sick    & 0.65 & 0.35 & 0.00 \\
              & Critical& 0.10 & 0.65 & 0.25 \\
        \bottomrule
    \end{tabular}
\end{table}

To supply expert knowledge for the Fuzzy-MAP EM algorithm, we simulated the construction of a fuzzy model based on domain expertise. Using the PyFume library \cite{fuchs2020pyfume}, we trained a Takagi-Sugeno fuzzy model on a separate, larger dataset, generated from the ground-truth POMDP, than the one used for POMDP learning. This approach ensured that the expert knowledge was derived independently from the target learning process. 
Functioning as an expert system, the fuzzy model takes the current observation $o_t$ and the selected action $a_t$ as inputs to generate a prediction of the next observation, denoted as $\hat{o}_{t+1}$.

The simulated expert model demonstrated satisfactory predictive accuracy, achieving $R^2$ scores of $0.63$ and $0.64$ for forecasting the next test result and symptoms profile, respectively, on a held-out test dataset.

The performance of the proposed approach is evaluated under two challenging conditions: (\emph{i}) a low-data regime, where learning is based on only 3 trajectories of length 5; and (\emph{ii}) a high-noise regime, where 10 trajectories of length 5 are perturbed by Gaussian noise $\mathcal{N}(0, 0.25)$ applied to each component of the observation vector.
Following a grid search sensitivity analysis, the hyperparameters for the experimental setup were selected as follows: $\lambda_T = 0.1$ and $\lambda_O = 0.05$.

\paragraph{Low-Data Regime} 
When data availability is extremely limited, the standard EM algorithm struggles to recover the underlying model, resulting in a high L1 distance for transition probabilities (0.43) and substantial KL divergences in the observation distributions (Table \ref{tab:results}, top part). In contrast, Fuzzy-MAP EM---guided by the expert model---produces a significantly more accurate transition model (L1 distance of 0.18) and yields more plausible observation distributions.

%\vspace{-0.5cm}
\paragraph{High-Noise Regime} 
In the high-noise scenario, the advantages of Fuzzy-MAP EM are more modest but still evident. It continues to learn a more accurate transition model (L1 distance of 0.26 vs.\ 0.29) and shows a marked improvement in modeling the observation distribution for the ``Critical'' state (Table \ref{tab:results}, bottom part).
%(Table \ref{tab:high_noise_results}).

\smallskip
Overall, these results demonstrate that incorporating expert fuzzy knowledge into the EM framework yields more robust parameter estimation under both data scarcity and observational uncertainty. 

%Additional experiments, employing alternative POMDP ground truths, are presented in the supplementary material, together with a sensitivity analysis of the model hyperparameters ($\lambda_T$ and $\lambda_O$).
%\todo[color=pink,textcolor=black]{One suggestion to save space: create a unique table to compare standard/fuzzy EM on low-data and high-noise scenarios. Just add one initial column titled 'Regime', and separe the results using a line. (DB)}
\begin{table}[ht]
    \centering
    \caption{Comparison of Standard EM and Fuzzy-MAP EM.}
    \label{tab:results}
    \begin{tabular}{l|c|c|c}
        \toprule
        \textbf{Regime} \hspace{0.30cm} & \textbf{Measure} &  \hspace{0.05cm} \textbf{Standard EM}  \hspace{0.05cm} & \hspace{0.05cm} \textbf{Fuzzy-MAP EM} \\
        \midrule
        \multirow{4}*{Low-Data} & \hspace{0.15cm} L1 distance transitions \hspace{0.15cm}  & 0.43 & \textbf{0.18} \\
        & KL divergence Healthy   & 3.43 & \textbf{0.19} \\
        & KL divergence Sick      & $\infty$ & \textbf{2.63} \\
        & KL divergence Critical  & 9.11 & \textbf{6.39} \\
        \midrule
        \multirow{4}*{High-Noise} & L1 distance transitions $\:$ & 0.29 & \textbf{0.26} \\
        & KL divergence Healthy   & 0.38 & \textbf{0.35} \\
        & KL divergence Sick      & \textbf{0.57} & 0.65 \\
        & KL divergence Critical  & 1.23 & \textbf{0.43} \\
        \bottomrule
    \end{tabular}
    \vspace{-\baselineskip}
\end{table}

%\begin{table}[ht]
%    \centering
%    \caption{Comparison of Standard EM and Fuzzy EM in the high-noise scenario.}
%    \label{tab:high_noise_results}
%    \begin{tabular}{l|c|c}
%        \toprule
%        Measure & $\:$ Standard EM $\:$ & $\:$ Fuzzy EM \\
%        \midrule
%        L1 distance transitions $\:$ & 0.29 & \textbf{0.26} \\
%        KL divergence Healthy   & 0.38 & \textbf{0.35} \\
%        KL divergence Sick      & \textbf{0.57} & 0.65 \\
%        KL divergence Critical  & 1.23 & \textbf{0.43} \\
%        \bottomrule
%    \end{tabular}
%\end{table}

\section{Therapeutic Protocols of Myasthenia Gravis}
\label{sec:miastehenia}
%\todo[color=pink, textcolor=black]{In blue, a new proposal for the title of this section. (DB)}

%spiegare in due parole MG.

%Motivare perchè utilizziamo solo i farmaci biologici

%\subsubsection{Fuzzy model of adaptive immune system alterations in MG}

In this section we assess the clinical relevance of our approach by applying the Fuzzy-MAP EM algorithm to a fuzzy model of Myasthenia Gravis (MG) defined by domain experts. The objective is to learn a two-state POMDP that captures the disease dynamics under two distinct scenarios: absence of treatment and administration of biological drug named Ravulizumab.
%, i.e., a complement-inhibiting therapeutic agent.

\subsection{Fuzzy Model of Adaptive Immune System Alterations in MG}

%\todo[color=pink, textcolor=black]{Shall we add a couple of sentences here to introduce MG and explain why it is a relevant real-case application (e.g. difficulty in identifying the optimal treatment)?\\
%I would also include a general bibliographic reference for this pathology. (DB)}
MG is a rare autoimmune disease that causes severe and debilitating muscle weakness.
The clinical manifestation is characterized by differences in age of onset, muscle involvement, auto-antibody specificity and thymic pathology, suggesting that multiple disregulated pathways can lead to the same disease.
Accordingly, a high diversity is also observed in the effect of therapies, as a substantial minority of patients are unresponsive to classical immunosuppressants.
The development of methods for personalized medicine based on patient's specific features is thus of particular interest.
For a comprehensive review of clinical and biological aspects of MG the reader can refer to \cite{Gilhus2019}.

\begin{figure}[htp]
    \centering
    \includegraphics[alt={Interaction map of the MG fuzzy model}, width=1.\linewidth]{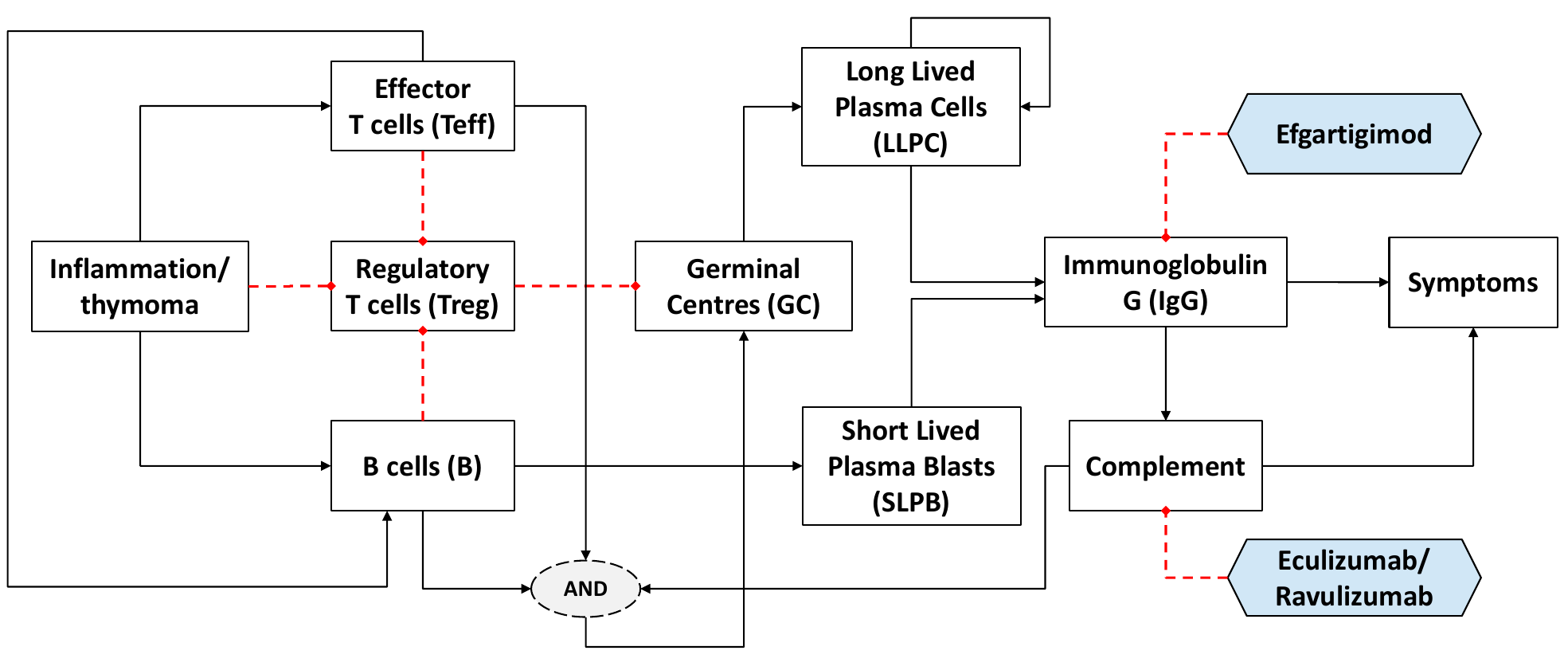}
    \caption{Interaction map of the MG fuzzy model. Plain black (dotted red) lines represent positive (negative) regulations. %dotted red lines represent negative regulations. 
    Coloured boxes correspond to biological drugs.}
    \label{fig:MGmodel}
\end{figure}

The fuzzy model considered in this work is focused on the most common type of MG---i.e. AChR-MG associated with thymic alterations \cite{Gilhus2019}---and takes into account the main cellular components involved in the associated autoimmune reaction, which is triggered by inflammation and/or thymoma (Fig. \ref{fig:MGmodel}).
These events cause alterations in the adaptive immune system, such as a decreased activity of regulatory T cells (Treg), and an increased activity of  autoreactive effector T cells (Teff) and B cells (B).
This autoimmune reaction results in the differentiation of autoreactive B cells into Short Lived Plasma Blasts (SLPB), which produce self-reacting Immunoglobulins G (IgG) that target and disrupt the neuromuscular junction, leading to the characteristic symptoms of MG.
IgG bound to biological surfaces can induce the activation of the complement system, causing the formation of opsonins (i.e., proteins that facilitate the phagocytosis of the bound antigen and elicit immune reactivity) and of the membrane attack complex (i.e., a protein complex that forms pores in the cells' membranes, destabilizing them).
The concurrent activity of self-reactive Teff and B cells and the presence of complement opsonized antigens in the hyperplastic thymus of MG patients, leads to the formation of ectopic Germinal Centres (GC), in which Plasma Cells (PC) producing high affinity antibodies are selected and proliferate. 
GC-derived PC are capable of establishing niches that sustain them for years, forming a population of Long Lived PC (LLPC) that continuously produce IgG, leading to long lasting and treatment resistant humoral autoimmunity\cite{cavalcante2013etiology}.

Biological drugs such as {\em Efgartigimod}, {\em Eculizumab} and {\em Ravulizumab} can be used in MG therapeutic protocols as final fallback to alleviate MG symptoms, especially for patients not properly responding to standard treatments.
The effectiveness of these drugs depends on their ability to specifically target molecules directly involved in MG pathology: 
Efgartigimod induces the decrease of IgG levels, while Eculizumab and Ravulizumab target the terminal complement component C5.
Other biological drugs targeting B cells and specific cytokines involved in autoimmune processes also exist \cite{vanoli2022antibody}.
The fuzzy rules and the linguistic terms of all model's variables are shown in Fig.\ref{fig:rules}, Appendix A.
%\vspace{-0.9cm}
\subsection{Experimental Design}
This case study aims to learn a two-state POMDP model designed to assess two alternative actions, i.e., {\em no intervention} and {\em treatment with Ravulizumab}, in the case where a 10-dimensional observation space is derived from the variables of an expert-defined fuzzy model.
Using the fuzzy model, a synthetic training dataset comprising 40 simulated patient trajectories was generated, with each trajectory consisting of 9 discrete timesteps.

To ensure a robust starting point and to mitigate convergence to poor local optima, the parameters were initialised using a data-driven approach. The k-means algorithm was used to partition the dataset into two clusters, whose centroids were used as the initial mean vectors for each state's observation model. To complete the setup, the initial transition probabilities were set to uniform, and the observation covariance matrices were initialised as identity matrices.

Hyperparameter selection was guided by a sensitivity analysis. Empirical observations revealed that assigning high values to $\lambda_T$ and $\lambda_O$ led the model to collapse into a single latent state.
For this experiment, we selected $\lambda_T = 0.05$ and $\lambda_O = 0.05$, as these values offered a balance between expert guidance and empirical validation. Upon completing the training, we performed a single iteration of the standard Expectation-Maximization (EM) algorithm to allow empirical data to modestly adjust the transition estimates, which might otherwise be overly constrained by the expert-informed fuzzy priors.

\subsection{Results}
\begin{figure}[b]
    \centering
    \includegraphics[alt={MG observation distributions},width=1.\linewidth]{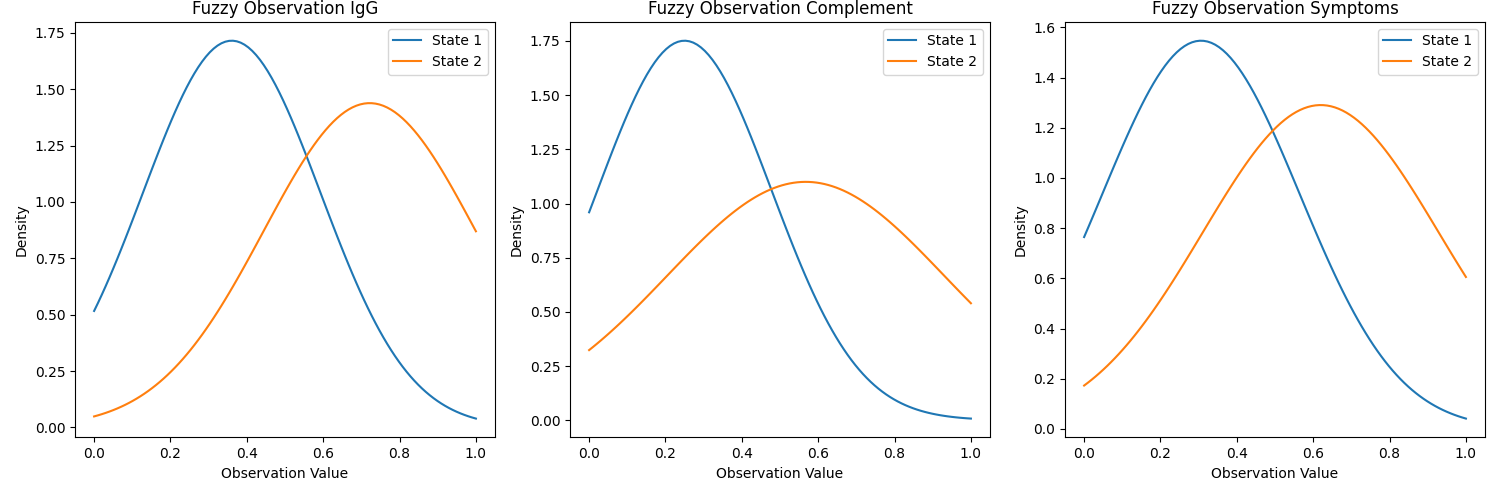}
    \caption{Learned observation probability density distributions for IgG, Complement and Symptoms variables, separated by latent state. State 1 (blue) represents the ``Mild MG" condition, and State 2 (orange) represents the ``Severe MG" condition.}
    \label{fig:mg_obs}
    \vspace{-\baselineskip}
\end{figure}

The application of  Fuzzy-MAP EM to MG yielded a robust two-state model. Based on the learned observation probability distributions, the latent states were semantically interpreted as \textit{Mild MG} (State 1) and \textit{Severe MG} (State 2). This interpretation is substantiated by the distributions of variables central to MG pathology.
Specifically, IgG and complement levels were significantly higher in the \textit{Severe MG} state, consistent with a greater autoantibody load and more pronounced complement activation (Fig. \ref{fig:mg_obs}). This clinical characterization is further supported by other model variables, wherein the \textit{Severe MG} state is associated with higher levels of symptoms, inflammation, and pro-inflammatory T cells (Teff). Conversely, the \textit{Mild MG} state reflects a more favourable immunological profile, characterized by reduced inflammation and symptoms severity.

The learned transition probabilities quantify both the natural disease progression and the therapeutic effect of Ravulizumab (Table \ref{tab:mg_transitions}). The administration of the drug shows a clear and positive clinical impact: for patients in the Severe state, Ravulizumab reduces the probability of remaining severe from 81.1\% down to 64.0\%.
This result successfully captures the drug's known mechanism of action confirming that, by inhibiting the complement system, it can alleviate the disease severity and promote a shift towards a healthier clinical state. Crucially, this modeled therapeutic effect is consistent with real-world clinical evidence, which  demonstrated the efficacy of Ravulizumab in treating MG patients \cite{vu2022terminal}. 

\begin{table}[b!]
\centering
\caption{Learned transition probabilities for the two-state MG model. \textbf{Bold} indicates a significant difference between the two actions.}
\label{tab:mg_transitions}
    \begin{tabular}{lccc}
        \toprule
        \textbf{Action} \hspace{0.8cm} & \hspace{0.25cm} \textbf{Current State} \hspace{0.35cm} & \multicolumn{2}{c}{\textbf{Next State}} \\
        \cmidrule(lr){3-4}
         & & Mild MG\hspace{0.4cm} & 
         \hspace{0.4cm} Severe MG \\
        \midrule
        No treatment & Mild MG & 0.979 & 0.021\\
             & Severe MG    & \textbf{0.189} & \textbf{0.811} \\
        \midrule
        Ravulizumab & Mild MG  & 0.968 & 0.032 \\
              & Severe MG    & \textbf{0.360} & \textbf{0.640}  \\
        \bottomrule
    \end{tabular}
    \vspace{-\baselineskip}
\end{table}

%\vspace{-0.2cm}
\section{Discussion and Conclusion}
\label{sec:conclusion}

The enhanced performance of our Fuzzy-MAP EM algorithm highlights the critical role of integrating expert knowledge into probabilistic modeling frameworks.
By leveraging \textit{fuzzy pseudo-counts} as an informative prior, the proposed method facilitates the learning process, where empirical evidence is regularised by domain expertise, resulting in more robust models when faced with limited or noisy data. This has significant practical applications in complex domains such as the management of rare diseases. The Myasthenia Gravis case study demonstrates this, illustrating how a reliable decision-making framework can be established even in the absence of initial patient data.

Nevertheless, our method presents certain trade-offs and limitations that are worth discussing. From a theoretical point of view, introducing a fuzzy prior does not guarantee the maximisation of the expected complete data log likelihood. Consequently, the monotonic increase of the data log-likelihood $\mathcal{L}(\theta)$ at each iteration is no longer guaranteed, which is a core property that supports the convergence proof for the standard EM algorithm. Secondly, the method is susceptible to errors originating from the fuzzy prior itself. We have identified two main issues:
\begin{itemize}
    \item \textbf{Prediction Bias:} inaccuracies in the expert fuzzy model are directly propagated into the learned POMDP parameters. %If the fuzzy model has a systematic bias (e.g., underestimating the severity of a state), this bias will be reflected in the learned observation distributions.
    \item \textbf{Antecedent Confusion:} this issue arises from the discrepancy between the fuzzy model's reliance on observations and the POMDP's latent state structure. When observation distributions from different states overlap, a rule originally \textit{intended} for one state (e.g., 'Healthy') can be erroneously activated by an observation from another (e.g., 'Sick'), causing the learned parameters to skew towards the state with the most confident prediction. 
\end{itemize}
These limitations thus highlight areas for future research.
%\todo[color=green]{Possible future work is to extend the paper with Type-2 Fuzzy models, analysis of the convergence, extend the case study with temporal dynamic for every variable(emivita).}

In summary, the present paper introduced the Fuzzy-MAP EM algorithm, a novel and original  method for integrating expert knowledge into the POMDP parameter learning process. The augmentation of the standard EM algorithm with \textit{fuzzy pseudo-counts} is a methodology that has been demonstrated to effectively mitigate the challenges posed by data scarcity and observational noise, thereby producing more accurate and robust models. As demonstrated through both synthetic experiments and a real-world case study on Myasthenia Gravis, this method represents a significant step toward developing reliable decision support systems in complex domains where data is inherently limited.

\begin{credits}
\subsubsection{\ackname}
This work has been developed within the research project: {\em “Personalising myasthenia gravis medicine: from “one-fits-all” to patient-specific immunosuppression”} (ERAPERMED2022-258, GA 779282), funded by Fondazione Regionale per la Ricerca Biomedica (Regione Lombardia), which also funded the PhD fellowship of ML and the research scholarship of RCC.
This work is also partially supported by the MUR under the grant ``Dipartimenti di Eccellenza 2023-2027'' of the Department of Informatics, Systems and Communication of the University of Milano-Bicocca, Milan, Italy, and by the National Plan for NRRP Complementary Investments (established with the decree-law May 6, 2021, n. 59, converted by law n. 101 of 2021) in the call for the funding of research initiatives for technologies and innovative trajectories in the health care sectors (Directorial Decree n. 931 of 06-06-2022)—project n. PNC0000003—AdvaNced Technologies for Human-centrEd Medicine (project acronym: ANTHEM). This work reflects only the authors’ views and opinions, neither the Ministry for University and Research nor the European Commission can be considered responsible for them.
\subsubsection{\discintname} No interests to declare.
\end{credits}

\bibliographystyle{splncs04}
\bibliography{bibliography}

\appendix

\counterwithin*{figure}{section}
\renewcommand{\thesection}{Appendix \Alph{section}.}
\renewcommand\thefigure{A.\arabic{figure}}   
\section{Fuzzy model of Myasthenia Gravis}
This section provides supplementary details on the expert-defined fuzzy model used to simulate the immunopathology of Myasthenia Gravis (MG).
%, as described in the main paper. 
The model is designed to capture the regulatory interactions among the key cellular and molecular components that drive the disease.

The conceptual structure of these interactions is illustrated in the interaction map in Figure \ref{fig:MGmodel} of the main manuscript. Figure \ref{fig:rules} below shows the model's variables (nodes), their linguistic terms, and the complete set of fuzzy rules that govern their behaviour.

\begin{figure}[htbp]
    \centering
    \includegraphics[width=1.05\textwidth]{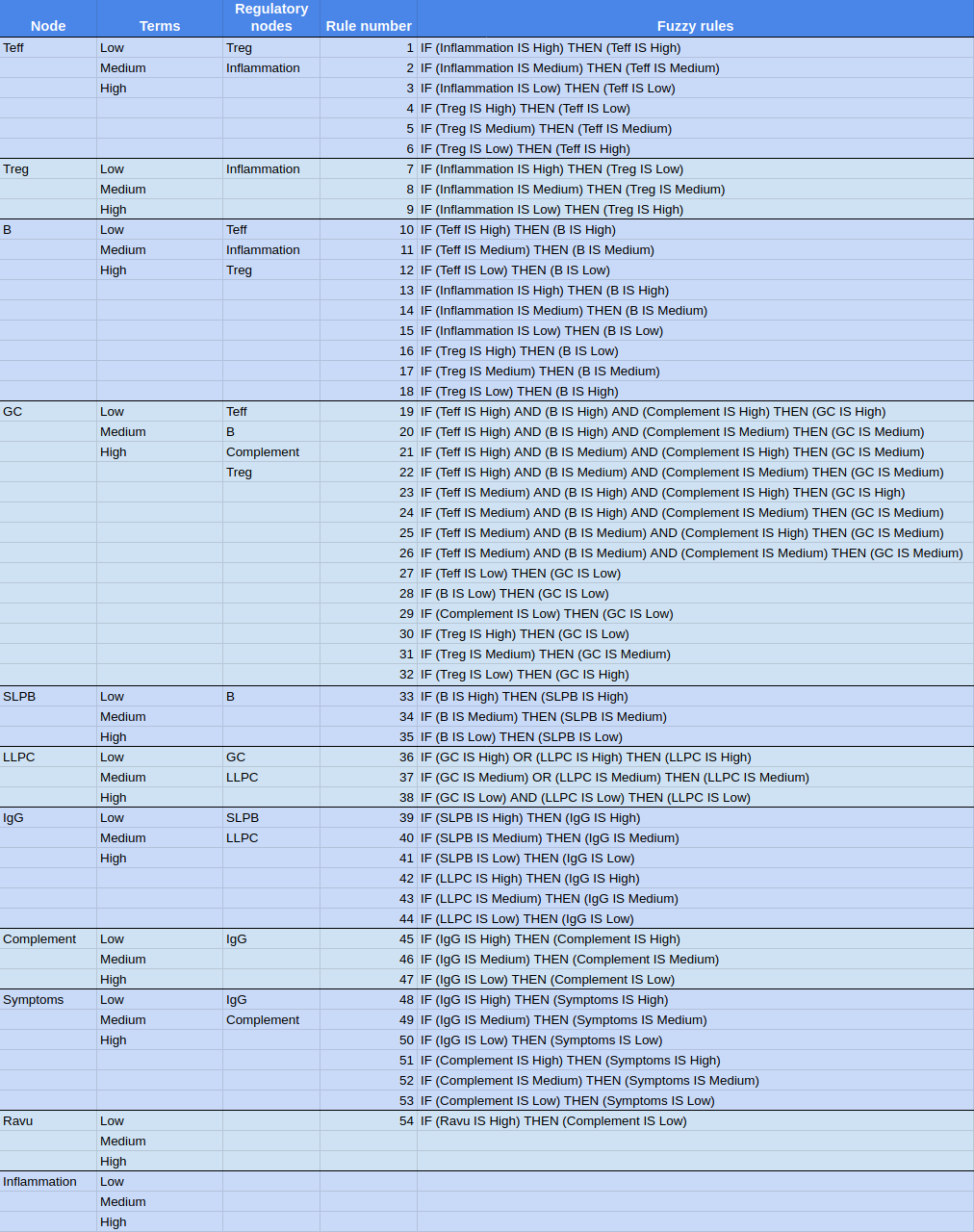}
    \caption{ Detailed specification of the fuzzy model for Myasthenia Gravis (MG) }
    \label{fig:rules}
\end{figure}
\end{document}